\documentclass[sigconf,screen]{acmart}
\AtBeginDocument{%
  }

\usepackage{amsmath,amssymb,amsfonts}
\usepackage{array}
\newcolumntype{C}[1]{>{\centering\arraybackslash}p{#1}}
\usepackage{enumitem}
\setlist[itemize]{noitemsep, topsep=0pt, parsep=0pt, partopsep=0pt}
\usepackage{algorithmic}
\usepackage{graphicx}
\usepackage{textcomp}
\usepackage{cleveref} 
\usepackage{booktabs}
\usepackage{todonotes}

\setcopyright{cc}
\setcctype{by}
\acmDOI{10.1145/3776734.3794427}
\acmYear{2026}
\copyrightyear{2026}
\acmISBN{979-8-4007-2321-6/2026/03}
\acmConference[HRI Companion '26]{Companion Proceedings of the 21st ACM/IEEE International Conference on Human-Robot Interaction}{March 16--19, 2026}{Edinburgh, Scotland, UK}
\acmBooktitle{Companion Proceedings of the 21st ACM/IEEE International Conference on Human-Robot Interaction (HRI Companion '26), March 16--19, 2026, Edinburgh, Scotland, UK}
\received{2025-12-08}
\received[accepted]{2026-01-12}

\begin{document}
\title{Design of a Robot-Assisted Chemical Dialysis System}

\settopmatter{authorsperrow=4}

\author{Diane Jung}
\orcid{0009-0004-8500-4162}
\affiliation{%
  \institution{University of Colorado at Boulder}
  \city{Boulder, CO}
  \country{USA}
}
\email{diju9182@colorado.edu}

\author{Caleb Escobedo}
\orcid{0000-0002-1616-956X}
\affiliation{%
  \institution{University of Colorado at Boulder}
  \city{Boulder, CO}
  \country{USA}
}
\email{caes5603@colorado.edu}

\author{Noah Liska}
\orcid{0009-0000-1142-8115}
\affiliation{%
  \institution{University of Colorado at Boulder}
  \city{Boulder, CO}
  \country{USA}
}
\email{noli3118@colorado.edu}

\author{Maitrey Gramopadhye}
\orcid{0009-0002-8184-9235}
\affiliation{%
  \institution{University of North Carolina at Chapel Hill}
  \city{Chapel Hill, NC}
  \country{USA}
}
\email{maitrey@cs.unc.edu}

\author{Daniel Szafir}
\orcid{0000-0003-1848-7884}
\affiliation{%
  \institution{University of North Carolina at Chapel Hill}
  \city{Chapel Hill, NC}
  \country{USA}
}
\email{daniel.szafir@cs.unc.edu}

\author{Alessandro Roncone}
\orcid{0000-0001-7385-1875}
\affiliation{%
  \institution{University of Colorado at Boulder}
  \city{Boulder, CO}
  \country{USA}
}
\email{alro6039@colorado.edu}

\author{Carson Bruns}
\orcid{0000-0002-4285-2725}
\affiliation{%
  \institution{University of Colorado at Boulder}
  \city{Boulder, CO}
  \country{USA}
}
\email{cabr6845@colorado.edu}

\renewcommand{\shortauthors}{D.N. Jung, C. Escobedo, N. Liska, M. Gramopadhye, D. Szafir, A. Roncone, and C.J. Bruns}

\begin{abstract}
Scientists perform diverse manual procedures that are tedious and laborious.
Such procedures are considered a bottleneck for modern experimental science, as they consume time and increase burdens in fields including material science and medicine.
We employ a user-centered approach to designing a robot-assisted system for dialysis, a common multi-day purification method used in polymer and protein synthesis.
Through two usability studies, we obtain participant feedback and revise design requirements to develop the final system that satisfies scientists' needs and has the potential for applications in other experimental workflows.
We anticipate that integration of this system into real synthesis procedures in a chemical wet lab will decrease workload on scientists during long experimental procedures and provide an effective approach to designing more systems that have the potential to accelerate scientific discovery and liberate scientists from tedious labor.
\end{abstract}


\maketitle

\section{Introduction}

\begin{figure}
    \centering
    \includegraphics[width=\linewidth, alt={A three panel image showing the experimental setup and two version of the dialysis membrane holder.}]{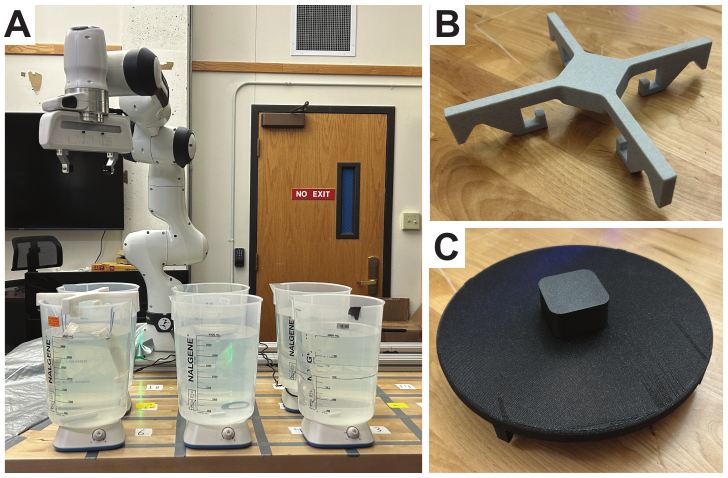}
    \Description{A three panel image showing the experimental setup and two version of the dialysis membrane holder.}
    \caption{Photo of the (A) experimental setup with the Franka Research 3 (FR3), six 4 L Nalgene containers filled with water, each placed on a stir plate. (B) Preliminary and (C) modified dialysis membrane holder prototypes.} 
    \label{fig:setup}
\end{figure}

Manual, laborious, and time-consuming tasks are performed routinely by experimental researchers, presenting a bottleneck for scientific progress and motivating the integration of robots and automation in laboratories \cite{angelopoulos2024transforming}.
In wet labs, most automation tools are designed either to perform specific tasks, or automate an entire experiment with minimal human intervention \cite{volk2023alphaflow,szymanski2023autonomous}. 
The latter high-automation cases demand computational and space resources, while relegating scientists to supportive tasks like system maintenance \cite{angelopoulos2024transforming}. 
The growing ubiquity of commercially available general-purpose robots presents an opportunity for human-robot collaboration to improve efficiency and safety in the wet lab. 
We investigate the application of a user-centered approach to design a practical, effective, and safe collaborative robotic system that will alleviate burdens of scientists working in the chemical wet lab.
This research was motivated by the challenge of the complex environment, high societal value of the application domain, and lack of prior human-robot interaction (HRI) research in this domain.

Based on previous research \cite{jung2024typology}, we selected dialysis as the first empirical case study to investigate the impact of human-robot collaboration for experimental tasks using a general-purpose collaborative robot.
Dialysis is a polymer \cite{neufeld1966use} and protein \cite{mcphie19714} purification method that is dull and inconvenient due to repetitive buffer-replacement tasks that occur in regular time intervals over multiple days \cite{andrew2001dialysis, berg2007biochemistry}.
Previously, dialysis automation was achieved through a flow dialysis system where fresh buffer solution was (i) continuously pumped into a closed vessel \cite{schuett2021dialysis} or (ii) filled and exchanged at predetermined times by a peristaltic pump \cite{terziouglu2022automated}. 
Both methods were integrated into a commercially available automated synthesis workstation.
While workstations have been used to fully automate experimental procedures, commercial robots working in collaboration with scientists offer a distinct advantage: the ability to operate safely alongside scientists in shared environments.
By leveraging capabilities such as collision anticipation \cite{escobedo2021contact}, object-aware control \cite{escobedo2022object}, and spill-free fluid transport \cite{abderezaei2024clutter}, these systems can adapt to the dynamic nature of a chemical wet lab.
We hypothesize that integrating these capabilities within a user-centered design process will ensure that the system effectively assists with dull and tedious tasks while laying the groundwork for broader human-robot teaming in experimental science.
In this work, we present a robot-assisted dialysis system as a pilot study to demonstrate the feasibility of this collaborative approach. We posit that this work lays the foundation for integrating general-purpose robots into experimental workflows, enabling safe, close-proximity assistance that relieves scientists from tedious labor.

\section{Background}
There has been a growing interest in studying how people interact with robots in various in-the-wild contexts, including airports \cite{nielsen2018subjective}, offices \cite{agrigoroaie2020wild}, hospitals \cite{eriksen2023understanding}, malls \cite{koike2025drives}, public streets \cite{pelikan2024encountering}, and homes \cite{ahtinen2023robocamp}. 
Other in-the-wild studies have focused on collaborative robotic assistance for labor-intensive and hazardous tasks in domains such as agriculture \cite{elias2025analyzing} and underwater archaeology \cite{jiang2025assistant}.
Furthermore, commercial robots have been deployed in conventional laboratory spaces for assistance with routine tasks.
Examples include research on optimal path planning for mobile robots in a life science lab \cite{liu2013mobile}, design of manipulators to handle lab equipment \cite{ali2016multiple}, development of safety requirements for robots in a lab \cite{fritzsche2007safe}, and design and implementation of a mobile robot in a life science lab \cite{kleine2022designing}. 
Gaps exist in development of human-robot collaborative systems in which a robot contributes directly to an experiment and investigation of how the system impacts the scientists. 

Integrating robots directly into an experimental procedure is challenging because labs are dynamic and space-constrained, often involving cross-traffic among multiple researchers completing various tasks. 
Moreover, high levels of automation can be detrimental to human performance with regard to mental workload, situation awareness (SA), complacency, and skill \cite{parasuraman2000model}.
Reduced SA also increases the time it takes the operator to intervene manually upon system failure \cite{endsley1995out, endsley1999level}.
Since these outcomes can potentially undermine research progress, they should be taken into account for the scientists' role and performance \cite{kaber2007human}. 
Additionally, past research suggests that low to moderate level of automation may increase SA \cite{kaber2004effects}, allowing scientists to remain engaged with the procedure, enabling better efficiency in error correction \cite{endsley1995out}, hands-on skill development \cite{parasuraman2000model}, and mentoring, maximizing enjoyment while still lowering the perceived workload associated with the task. 

Considering these complexities, we aimed to implement a human-in-the-loop workflow, where scientists customize experimental parameters and continuously monitor the experimental process, while the robot completes the dull and repetitive tasks.
To ensure we incorporate scientists' needs, we adapted a user-centered design approach, which involves users early in the iterative design process \cite{ISOdef}.
We hypothesize that early and frequent user engagement will also result in increased acceptance upon real-world deployment and seamless integration into real synthesis procedures with minimal disruptions to the scientists' original workflow.

\section{Methods}
In this study, we developed a robot-assisted dialysis system through a user-centered design approach. 
This protocol was reviewed and approved by the Institutional Review Board (Protocol \#24-0274).

\paragraph{Participant Recruitment.}
We initially recruited three research scientists through purposeful sampling and expanded the participant group size to five through snowball sampling \cite{patton2014qualitative}. 

\paragraph{Preliminary Interviews.}
We conducted preliminary semi-struc\-tured interviews with all participants to learn about various experimental parameters, procedures, and overall experiences. 
We utilized this information to design prototypes that satisfy the shared needs of wide-ranging dialysis procedures. 

\paragraph{User Feedback Sessions.}

All user feedback sessions were conducted in a dry lab environment with two dialysis membranes submerged in 4 L Nalgene containers filled with water. 
We began by introducing participants to the system and providing time to review the written instructions and familiarize themselves with system components.
We used interactive think aloud (ITA) \cite{patton2014qualitative, o2023talking}, to understand participants' thought processes and concerns. 
In our case, interactions were limited to (i) asking for explanations on decisions or nonverbal cues, (ii) answering participants' questions about the system, and (iii) asking participants to use both kinesthetic and gamepad guidance to save container locations. 
Afterwards, participants shared their opinions and suggestions.
To encourage active participation in the design process, we engaged in a brief solution ideation with the participants, where we discussed ways of addressing the needs specific to the participants' workflow.

The preliminary interviews, feedback sessions, and follow-up discussions were voice recorded and transcribed. 
We conducted thematic analysis \cite{braun2006using, saldana2021coding} to identify common themes associated with confusion and suggestions. 
Finally, the research team reviewed the themes presented by two or more participants, discussed additional refinements based on our observations during the sessions, and finalized a list of system requirements for subsequent versions.

\section{System Prototype}

\begin{figure*}
    \centering
    \includegraphics[width=.92\linewidth, alt={Images of the graphical user interface.}]{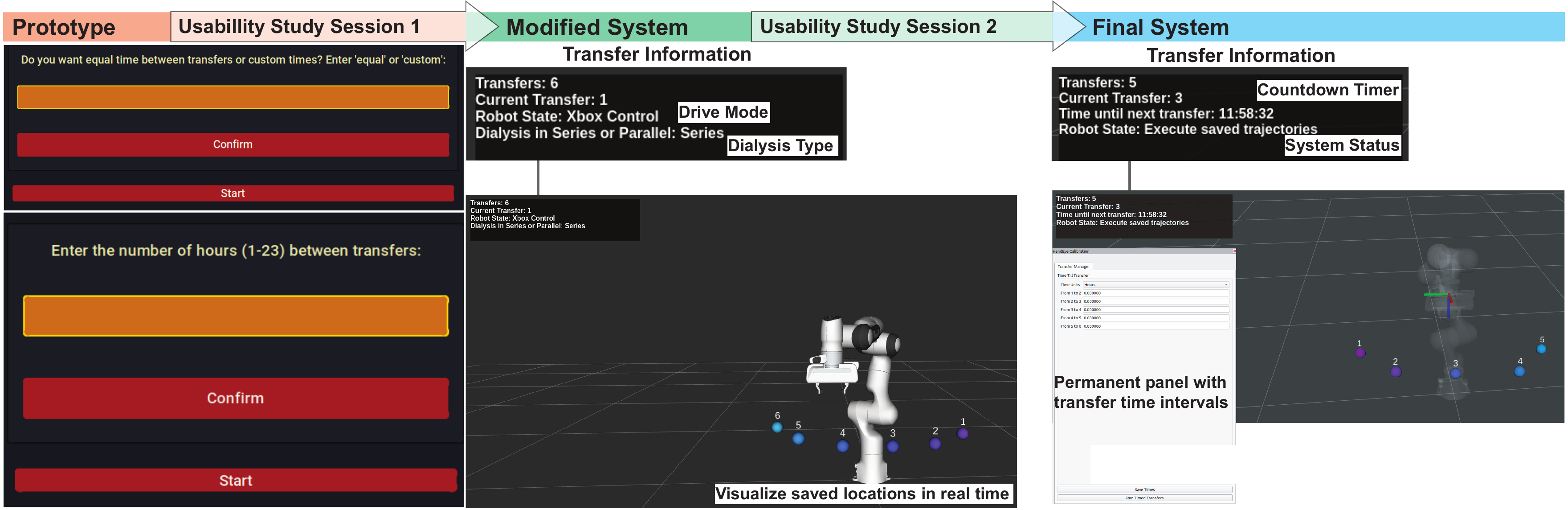}
    \Description{Images of the graphical user interface.}
    \caption{Graphical user interface of the robot-assisted dialysis system with key improvements emphasized in white boxes.}
    \label{fig:robodialysis_PDsummary}
\end{figure*}

We aimed to design a stand-alone robot-assisted dialysis system, independent of peripheral laboratory equipment, while preserving as much of the standard protocol as possible. 
Typically, dialysis occurs over 3--21 consecutive days, with several liters of buffer solution being exchanged every 3--24 h. 
Informed by these protocols, we prototyped a robot-assisted dialysis system (Fig. \ref{fig:setup}A) using a Franka Research 3 (FR3) robotic arm based on two initial system requirements: (i) the system will consist of a maximum of six containers, each filled with 4 L of buffer solution, placed on stir plates and (ii) the dialysis membranes will be transferred from one container to the next by the FR3 at user-defined time intervals. 
In our system, scientists have full control over dialysis membrane preparation.

\paragraph{Dialysis Membrane Holder.}

To support the robotic process of moving all dialysis membranes from one container to the next, we designed a custom 3D-printed dialysis membrane holder to sit on the edge of 4 L Nalgene containers. 
The preliminary design (Fig. \ref{fig:setup}B) of this membrane holder was X-shaped, where four arms extend from a rectangular center to establish maximum surface contact between the membrane holder and the robot end effector, enabling secure grasps. 
Each arm consists of two key components: (i) a hook, where dialysis membranes are suspended by a cable tie or rubber band, and (ii) a trapezoidal cut to compensate for positioning errors and thus ensure the membrane holder is centered on the container.

\paragraph{System Setup.}
Once the prepared dialysis membranes are suspended on the membrane holder, participants fill up to six containers with buffer solution.
While manual dialysis procedures typically involve emptying and refilling the same container with fresh buffer solution at designated time intervals, we modified the process to begin with pre-filled containers so that participants can avoid manual buffer-refilling operations throughout the multi-day procedure. 

\paragraph{Save Container Locations.}
Next, participants customize when and how to transfer the membrane holder between containers.
Setting the sequence of container locations is achieved by kinesthetic or gamepad guidance, labeled as ``hand guide mode'' and ``Xbox controller mode'' in the user instructions, respectively.
Participants thus guide the FR3 using their preferred method to grip the dialysis membrane holder with the robot end effector and use the gamepad to save the locations of each container in sequence.

\paragraph{Graphical User Interface (GUI)}
Our preliminary GUI (Fig. \ref{fig:robodialysis_PDsummary}: Prototype) consists of two customizable parameters: (i) ``equal'' or ``custom'' transfer time intervals and (ii) time between transfers. ``Equal'' times are for procedures where transfers occur at fixed time intervals and ``custom'' times are for procedures where the transfer times change throughout the procedure. 
Once participants finish saving all container locations and customizing transfer times, they execute the automated transfers by clicking the start button on the GUI and pressing a button the gamepad. 
The robotic arm stays dormant between the transfers and moves only at the designated times to transfer the membrane holder.

\paragraph{Level of Automation.}
Using the Levels of Robot Autonomy \\(LORA) and the Sense–Plan–Act framework \cite{beer2014toward}, we employ a low to moderate autonomy level, LORA 3–4.
Participants manage the sense phase, providing task context and confirming contact conditions, and the plan phase by specifying grasp locations through kinesthetic or gamepad guidance and setting the transfer schedule.
The act phase is automated as the robot executes the membrane transfers at the specified times but does not generate, choose, or revise plans.
Additionally, the robot monitors force/torque thresholds as a safety check but does not perform autonomous perception.

\paragraph{Testing and Validation.}
We validated system consistency and precision in a dry lab environment.
We prepared mock dialysis membranes, suspended them on the membrane holder using rubber bands, and completed multiple transfers between 2--3 4 L Nalgene containers, each placed on a stir plate. 
In 12 h intervals, the FR3 grasped the membrane holder and moved it from one container to the next. 
After system validation, we developed a written list of instructions to guide users through system setup.

\section{Results}
We obtained insight on how scientists interact with the system, their preferences, needs, and points of confusion through two usability studies.
We prepared 2--3 4 L Nalgene containers, filled with 4 L of water, and two dialysis membranes filled with 1:1 tomato soup:water mixture, which were suspended on the membrane holder using rubber bands. 
Usability studies were conducted in a dry lab.

\begin{table*}
\centering
\caption{Summary of the first user feedback sessions.}
\label{table:req1}
\begin{tabular}{ p{8.6cm}|p{8.2cm}}
\toprule
\multicolumn{1}{c|}{\textbf{Participant Comments (number of users)}} & \multicolumn{1}{c}{\textbf{Solutions Implemented in the Modified System}} \\ 
\midrule
Multiple people conduct dialysis simultaneously in the lab (4). & Add ``parallel'' (simultaneous) and ``series'' (sequential) dialysis. \\
\midrule
Uncertainty about saved container locations (2). & GUI will show saved container locations and drive mode. \\
Request to see transfer progress (2). &  \\
\midrule
Confusion on pressing two buttons to execute (2). & One gamepad button will execute transfers. \\
\midrule
The robot moves too fast during gamepad control (2).  & All movements will be slower. \\
\midrule
Request for container lids (1). & [Addressed in final system (Table. \ref{table:req2})] \\
\midrule
Whether saving state is independent of drive mode (2) and grip (2). & Add clarifications on the written instructions. \\
Reminder to move end effector away from containers (2). &  \\
\midrule
Confusion on ``equal'' and ``custom'' times (2). & Remove function. \\
\bottomrule
\end{tabular}
\end{table*}

\begin{table*}
\centering
\caption{Summary of the second user feedback sessions.}
\label{table:req2}
\begin{tabular}{ p{9.2cm}|p{7.6cm}}
\toprule
\multicolumn{1}{c|}{\textbf{Participant Comments (number of users)}} & \multicolumn{1}{c}{\textbf{Solutions Implemented in the Final System}} \\ 
\midrule
Request for more information about transfer progress (3). & GUI will show transfer times, robot state, countdown timer.\\
\midrule
Request for container lids (1). Confusion on end effector positioning (2). & A cap will be added to the membrane holder.\\
\midrule
Concerns about longer membranes making physical contact with the container (2). & Pick-up height will be 2.2x the container height to accommodate all membrane lengths.\\
\bottomrule
\end{tabular}
\end{table*}

\paragraph{First Session.}

During the first session, we focused on investigating participants' preference between kinesthetic and gamepad guidance methods.
We identified ten themes referenced by two or more participants and developed six requirements for the next iteration (Table. \ref{table:req1}). 
Four participants suggested a new feature that would enable ``parallel'' dialysis to accommodate simultaneous procedures. 
Each of the following points were articulated by two participants: (i) confusion on whether a location had been saved, (ii) request to see more information about transfers, (iii) confusion on pressing a gamepad button and a GUI button to start transfers, (iv) fast movement of the robot during gamepad control, and (v) confusion on whether drive mode or grip status affects the function to save locations.
Additionally, two participants asked questions about ``equal'' and ``custom'' times, one participant suggested container covers, and the research team reminded two participants to move the end effector away from containers before starting the transfers.

Based on these comments, we developed the following six system requirements for the second prototype (Fig. \ref{fig:robodialysis_PDsummary}: Modified System): (i) add options for ``parallel'' and ``series'' dialysis, (ii) visualize saved container locations and indicate drive mode on the GUI, (iii) initiate the transfers with a press of one button on the gamepad, (iv) decrease the speed of robot movement during gamepad guidance, (v) add more details on the written instructions, and (vi) remove the wording, ``equal'' and ``custom'' times (Table. \ref{table:req1}). 
Additionally, we observed some participants saving container locations without the dialysis membrane holder and others saving locations with the holder. 
To avoid associated errors, we stressed that all locations must be saved with the membrane holder in the instructions.

\paragraph{Second Session.}

We focused on investigating usability and perception of changes. 
All five participants felt that gamepad control was easier and we received many positive comments on the new GUI, which provided real-time visualization of container locations, robot position, and robot movement in rendered 3D space.

We developed three system requirements based on four themes referenced by two or more participants (Table. \ref{table:req2}). 
Three participants suggested more real-time information about the transfers, two participants were concerned about the low lift height, which risks collision between the bottom of the suspended membranes and the top edges of the containers.
Two participants sought confirmation on end effector height positioning to avoid either a missed grasp (too high) or contact with the buffer solution (too low). 
Additionally, a second participant suggested a container cover.

Based on user feedback, we modified the requirements for the final system (Fig. \ref{fig:robodialysis_PDsummary}: Final System): (i) add a status panel on the GUI to display a countdown timer, transfer time intervals, and robot state, (ii) modify the dialysis membrane holder to provide a full covering over the containers (Fig. \ref{fig:setup}C), and (iii) set the lift height to 2.2x the height of the containers to accommodate all membrane lengths (Table. \ref{table:req2}). 

\section{Conclusion}

We employed a user-centered approach to design a human-in-the-loop dialysis system, based on a commercial robotic arm operating at low-to-moderate autonomy to leverage human dexterity, preserve the opportunity to practice hands-on skills, and enable easy manual interventions when necessary. 
Through two usability studies, we discovered potential operational complexities and distinct ways users interact with the system.
Based on participant feedback, we iterated on the system design to increase usability, applicability, and functionality.
In future work, we will integrate our system into real synthesis procedures conducted in a wet lab to evaluate effectiveness in purifying products and impact on scientists.

\section{Acknowledgments}

This work was funded by the National Science Foundation under Grant No. 2222952.
We acknowledge Dusty Woods for assistance with manufacturing the dialysis membrane holder.

\balance
\bibliographystyle{ACM-Reference-Format}
\bibliography{bibliography}

\end{document}